# Why are LLMs' abilities emergent?

Vladimír Havlík[1]

**Abstract**

The remarkable success of Large Language Models (LLMs) in generative tasks has raised fundamental questions about the nature of their acquired capabilities, which often appear to emerge unexpectedly without explicit training. This paper examines the emergent properties of Deep Neural Networks (DNNs) through both theoretical analysis and empirical observation, addressing the epistemological challenge of "creation without understanding" that characterises contemporary AI development. We explore how the neural approach's reliance on nonlinear, stochastic processes fundamentally differs from symbolic computational paradigms, creating systems whose macro-level behaviours cannot be analytically derived from micro-level neuron activities. Through analysis of scaling laws, grokking phenomena, and phase transitions in model capabilities, I demonstrate that emergent abilities arise from the complex dynamics of highly sensitive nonlinear systems rather than simply from parameter scaling alone. My investigation reveals that current debates over metrics, pre-training loss thresholds, and in-context learning miss the fundamental ontological nature of emergence in DNNs. I argue that these systems exhibit genuine emergent properties analogous to those found in other complex natural phenomena, where systemic capabilities emerge from cooperative interactions among simple components without being reducible to their individual behaviours. The paper concludes that understanding LLM capabilities requires recognising DNNs as a new domain of complex dynamical systems governed by universal principles of emergence, similar to those operating in physics, chemistry, and biology. This perspective shifts the focus from purely phenomenological definitions of emergence to understanding the internal dynamic transformations that enable these systems to acquire capabilities that transcend their individual components.

**Introduction**

Recent advances in Deep Neural Networks (DNNs) and their application to Large Language Models (LLMs) have led to the surprising success of these models in their generative capabilities, particularly in language translation, comprehension, and natural language dialogue, as well as in the audio-visual domains, where they generate images, music, and video with similar success. In general, these advances are the result of efforts to create artificial intelligence (AI) or artificial general intelligence (AGI), which have emerged in conjunction with the development of computers. Recent breakthroughs have been made possible primarily through the development of the required deep learning technologies (DL) for training DNNs, especially sufficient processor performance and sufficient digitised data. The empirical fact of the existence of these sophisticated abilities necessitates an explanation for these acquired abilities, which in many cases are indistinguishable from those previously attributed only to humans. A remarkable and surprising fact is that we can utilise these generative artificial intelligence technologies without fully understanding *how and why they work*. One of the hypotheses that has emerged from research into the capabilities of individual DNNs is that the complexity of behaviour in a network of neurons leads, under specific conditions, to emergent phenomena that result, usually suddenly or by leaps and bounds, in the behavioural capabilities of the model. Sufficient *complexity* is often a prerequisite in other areas of science, such as physics, chemistry, biology and economics, for phenomena that are frequently referred to as 'emergent' to occur. This raises the question of whether DNNs are a new area for research into these processes and whether it is justifiable to claim that the phenomena that occur and the abilities that models are able to acquire are emergent.

---

[1] Institute of Philosophy, Czech Academy of Sciences, Prague and University of West Bohemia, Pilsen, Czech Republic. Email: havlik@flu.cas.cz; ORCID: https://orcid.org/0000-0003-3529-614X

An affirmative answer to this question is supported by the fact that existing LLMs are now so complex that we are forced to study them as special natural phenomena. Although networks and the interactions between their components are a standard part of complexity theory(Jensen 2022), the specificity of the behaviour of LLMs as they acquire new capabilities has not yet been unequivocally supported by more fundamental theory. Thus, there is still a need to experiment and interpret results to orient us among the hypotheses presented. For some, this situation is comparable to physics in the early 20th century, as Boaz Barak commented: "We have a lot of experimental results that we don't completely understand, and often when you do an experiment it surprises you." (Heaven 2024) Thus, possible answers to the question of the emergence of these abilities in LLMs are produced by empirical observations in experiments, theoretical assumptions drawn from analogies in other areas of science, such as physics or complexity theory, and finally, conceptual philosophical discussions.

Ongoing discussions on the emergent abilities of DNNs have two main motivations: (1) to explain how LLMs in general acquire emergent abilities, how they move from mere memorization to generalizations, whether these are cases for which they are directly trained or even those for which they are not directly trained in the learning process; (2) to explain upon what the ability to acquire these abilities depends and whether they can be predicted based on some causal parameters. Since the discussions cannot yet be based on the generally accepted status of emergent phenomena and their eventual definition, it is often necessary to refer only to empirically established facts and to generalise hypothetical conclusions from them. Thus, the aim of this paper is to consider certain aspects of current generative AI technologies that entitle us to assume that the properties generally inherent in DNNs have the status of emergent phenomena, and what is relevant to their possible exploration and prediction based on this fact. The paper is organised as follows: the first three sections discuss some of the essential theoretical assumptions and contexts for the creation of generative AI, including the difference between symbolic and neural approaches, the question of determinism versus indeterminism, and the role of stochasticity and nonlinearity in these systems. The next two sections consider empirically discovered and unexpected phenomena such as scaling, grokking and delayed generalisation that are associated with the emergence of LLMs' capabilities. The final section assesses emergent properties in DNNs.

1. **Creation without understanding**

I have already mentioned the complexity of existing DNN models and the fact that these phenomena need to be investigated as a new area of special natural phenomena. In this context, I would like to emphasize two fundamental and, thus far, little-emphasized epistemological aspects of this situation. (1) Research on these new phenomena is not surprising in itself; what may be surprising is that the phenomena we are trying to understand and explain have not been revealed as existing independently of us, but that we have, in a literal sense, created them without the need for full knowledge of how and why they work as they do. (2) If it is possible to create such kinds of phenomena that we do not fully understand, how are these phenomena possible, and what do they govern if we do not govern them? While this fact raises many pressing existential, ethical, economic, and social questions, in this paper I will focus solely on the epistemological realm and demonstrate how something like *creation without understanding* is possible. Strange as it may seem, let us assume that this is a consequence of the emergent behaviour of DNNs, which are idealised models of natural neural structures. In general, emergent phenomena have the status of "brute empirical facts", where higher qualities emerge from the lower level of existence in which they are grounded, but do not belong to it, because they constitute a new order of existence and its special laws of behaviour. Conceptual problems of emergence are thus linked to the mutually exclusive requirements of dependence and independence, where emergent phenomena are in some sense dependent on their basal level, but at the same time are in some sense independent of it and autonomous in themselves (Havlík 2022).

It should be emphasised that the phrase *creation without understanding* is somewhat exaggerated and cannot be taken completely literally. The creation of DNNs was not an unintended result of our activity, but a targeted effort to create such models. The aforementioned lack of understanding is linked to certain reductionist assumptions that cannot be fulfilled. What is crucial in this case is the simplicity and complexity of the phenomena with which we are dealing. The functionality of the biological neuron is well understood and serves as the starting point for its idealised artificial model.[2] Both biological and artificial neurons perform essentially simple operations and behave strictly according to simple mathematical rules. The neuron receives signals at its inputs, performs a weighted summation of the signals, and outputs a signal based on a threshold function. Thus, a biological or artificial neuron does nothing surprising or mysterious. The first mathematical model of the nervous system was proposed by Warren McCulloch and Walter Pitts as early as 1943, as a network of simple logic elements (neurons). By connecting them together, they demonstrated that their theoretical model could perform all logical functions(McCulloch and Pitts 1943). Such a network model did not yet require or even contemplate "learning", nor could it be practically implemented given the technology of the time. However, the idea of "learning" was similarly derived from brain research, when in the late 1940s D. O. Hebb (2002 [1949]) proposed a hypothesis based on the mechanism of neural plasticity (called Hebbian learning), which was later used in many early neural network models, such as the Rosenblatt perceptron (1958) and the recurrent Hopfield network (1982).

Although the basic concept of entities (neurons) and their functioning in networks is well known, based on the study of small elementary models, it is no longer easy or even practically possible to predict what abilities will emerge as the parameters of these models increase. Thus, with only a certain amount of exaggeration, it can be said that no one knows exactly *how and why a given model works*. As researchers at AI company Anthropic, for example, say, "We understand exactly the mathematics of the trained network—each neuron in the neural network performs simple arithmetic operations—but we don't understand why those mathematical operations lead to the behaviors we see." (Anthropic 2023b)

It must be said that this is not the first time we have encountered such a situation in science. All emergent phenomena in different areas of science behave similarly and can be characterised more generally due to certain common features of their behaviour. Thus, if we do not understand why a deep neural network suddenly acquires the ability to solve a certain type of mathematical problem, then by analogy with similar situations in other areas of science, we can guess that the reason may be the manifestation of the contributions of individual neurons in the network, which become established in the systemic behaviour of the whole. Thus, while we can still describe the behaviour of each individual neuron, we cannot simply infer the systemic consequences of all such individual contributions. The fact that they occur is an empirical fact. They are caused by cooperative contributions from individual neurons in the network but due to its complexity it is practically impossible to predict when the network will reach a state in which such a capability will manifest itself. "Despite the ever increasing number of benchmarks that measure task performance, we lack explanations of what behaviors language models exhibit that allow them to complete these tasks in the first place." (Holtzman et al. 2023) Herein lies the success and difficulties of the neural approach to AI.

---

[2] The model of a neuron is idealised because it models its basic information transfer functionality but not its full biological behaviour. A biological neuron is often itself rhythmically active and drives activity for other connected neurons, which may also inhibit it. Such biological neuron activities are natural but are not modelled in DNNs.

## 2. Symbolic and neural approaches

To illustrate in more detail why the existence of emergent properties is likely within the neural approach, let us briefly recall how the neural approach differs from the traditional symbolic approach. The symbolic approach, unlike the neural approach, requires the formulation of sequentially ordered logical rules—algorithms—whose execution leads to desired outcome states, such as finding the value of a function or, more generally, processing data from inputs to outputs. Although the processing of such algorithms is seemingly instantaneous and can be executed using parallel processes, each rule of the algorithm is processed independently, and by tracing the execution of the algorithm rule by rule, we can continuously control the values taken by the variables, and the determinism of the execution of such an algorithm is fully predictable. The theoretical model for such an approach to an algorithm is the Turing machine (TM),[3] which represents the classical basis of computational theory, in which computation follows certain, predictable paths: each state and input symbol has a defined action, the same initial values and inputs lead to the same sequence of steps, and the result is always the same for a given input.

The neural approach utilises the neural network model mentioned above, which is derived from the functioning of neurons in the brain. The behaviour of the network is not programmed by sequential rules as in the symbolic approach, but the network has to go through a deep learning (DL) phase, where it is trained on the data so that over many epochs the weights are gradually adjusted (e.g. by a backpropagation mechanism) at each node of the network depending on the relationship between the input and output data. For example, if a model learns to recognise the form of a cat based on many different cat representations, this does not mean that the model will only remember all the cat representations available in the data during deep learning; the model will be able to recognise the cat even in representations it has never seen. It will learn to recognise a sort of general and characteristic "cat" pattern that it can apply to recognise cats it has not yet encountered. The model thus acquires, through learning, the ability to "recognise cats", which is a property of the model as a whole and not of a single neuron or a single layer. While each recognised representation of a cat is a given vector that has been evaluated by the model on the output layer of neurons as a "cat", the ability to recognise and evaluate such representations is itself spread across the model's weights. If this ability is not a direct consequence of one identifiable part but of the whole, then there is reason to view it as an emergent system property of the whole.

---

[3] There are deterministic and non-deterministic forms of TM; however, it is not essential to the difference between TM and DNNs. The difference lies not in determinism/indeterminism, but in the step-by-step traceability of TM's algorithms and the "black box" non-interpretability of DNN models.

## 3. Indeterminism, non-linearity and stochasticity

The true indeterminism of DNN behaviour is only apparent, and there are many essential sources for this false intuition. In contrast to the step-by-step traceability of a Turing machine, which clearly distinguishes the correctness of an algorithm's execution from its failure, a neural network must deal with "indeterministic" resources provided by both the model and its training and data (Antonion et al. 2024). Thus, random elements are crucial in both the deep learning phase and in the architecture of the model itself, as well as in its implementation. Simply put, it is thus impossible, for example, to deterministically predetermine whether the model will recognise the next image of a cat as "cat". This is because there are no algorithmic rules by which we can verify this in advance, as in the case of a Turing machine. Thus, cat recognition is indeed an acquired ability of the model but it was acquired in a specific way on specific data and is now hidden in the model among the many network parameters (weights, biases, number of neurons and layers and their architecture) and such implicit and distributed knowledge is not transparent or easily traceable and generally it is not possible to transpose it into a symbolic representation using logical rules. Therefore, it is not entirely clear or identifiable what general patterns or characteristics of the cat have been adopted by a given model and how exactly these patterns are stored in the millions or billions (and ultimately, trillions) of network parameters. Only the behaviour of the model, i.e., its percentage success in recognising cats, is evident. Thus, we should distinguish between the model's macro behaviour, which is recognisable, and its internal micro-behaviour, which, on the other hand, is difficult to interpret. These micro and macro perspectives are particularly important in the discussion of the emergent abilities of models.

From only the input-output point of view, a TM and a DNN are both deterministic systems. In theory, a trained neural network with fixed weights is a deterministic system that produces the same output for a given input. However, such strict determinism would be uninteresting in light of the natural diversity and creativity of language and other audiovisual outputs. Stochasticity, therefore, plays a fundamental role in the functioning of AI models.

First of all, there are many deliberately implemented stochastic software features, i.e. model parameters such as "temperature", "top-p", "top-k", which lead the model to generate results more randomly and thus more "creatively". Many neural networks, especially generative models such as LLMs, produce probability distributions and use parameters to control the randomness of these outputs. The stochastic behaviour of the models also affects the deep learning process and its optimisation, as the weights of the neural connections are initialised with random values, gradient descent and its variants are randomly applied, training data are also randomly selected and sampled, and other control techniques and optimisation algorithms that incorporate stochastic elements are used.

Second, the resulting behaviour of the model may also be influenced by various unintentional hardware and software causes. For example, the behaviour of models is heavily dependent on floating-point operations, which introduce very small internal random variations in the rounding of decimals, but which have very different consequences in the overall behaviour and properties of the models. Thus, this is primarily a given technical implementation of the hardware that affects these operations, such as the implementation of CUDA operations and multithreading (Xiao et al. 2021). It would seem that very small internal random deviations in the rounding of decimal places should not have a decisive influence on generative outputs. However, it is essential that models are highly sensitive to the initial values of weights, and any seemingly negligible change in the weight value of a given neuron can significantly affect the behaviour of the entire network. Thus, for example, empirical studies have found that even "changes of the order of $10^{-10}$ in a single weight at initialization can have the same effect as reinitializing all weights to completely random values"(Summers and Dinneen 2021). The finding of such sensitivity to initial values was surprising and is a consequence of the fact that a neural network (with at least one hidden layer of neurons) "is a nonlinear function parameterized by the model parameters θ (i.e., the network weights) that maps a measurable input set A to a measurable output set B." (Antonion et al. 2024) Similar to other dynamical systems whose behaviour we refer to as deterministic chaos (e.g., weather), this does not imply that such systems are unpredictable and indeterministic; rather, they are extremely sensitive to initial conditions. Edward Lorenz notably expressed this as the "butterfly effect" (Lorenz 1972; 2008), when he found sensitivity of his weather models to neglecting values from orders of $10^{-6}$ to $10^{-3}$, while in the above case sensitivity to a distribution of weights in the network of orders of $10^{-10}$ was observed.

In some cases, discussions have already begun about whether the actual mechanisms by which neural networks function are analogous to the behaviour of dynamical systems in chaotic theory (Kent 2023; S. Liu et al. 2024; Li et al. 2025). This idea is accepted but not yet elaborated in detail. Chaos, in this case, presents complex dynamics arising from nonlinearity and high sensitivity to initial conditions, as illustrated in the above examples. This is another crucial sign of general complex systems, suggesting a search for underlying principles that govern complex information processing, whether in natural or artificial systems. Furthermore, the analogy appears plausible at first glance, given that both chaotic systems and DNNs are characterised by high dimensionality and inherent non-linearity – factors known to be prerequisites for complex and sensitive behaviour and emergence.

Neural networks, as nonlinear stochastic systems, utilise nonlinear activation functions (e.g., ReLU, Sigmoid, Tanh) as fundamental components of the models. These nonlinear functions allow networks to learn complex patterns and representations that would be unattainable for linear models. Thus, nonlinearity is not only the reason for the high sensitivity of behaviour in DNNs, but is primarily a prerequisite for the ability of models to learn complex behavioural patterns. In fact, this combination of nonlinearity and randomness is beneficial to the performance of DNNs as it enables them to avoid local minima during training and thus explore more diverse solution spaces, contributing to their remarkable capabilities in solving complex tasks such as pattern recognition and generation. If the individual hidden neural network layers were linear, it would not matter how many hidden layers contributed to the outcome within the model, as linearity would allow all hidden layers to be merged into one. Such a model would be unable to solve complex and intricate questions and problems. It is the composition of multiple separate hidden layers that allows one to create nonlinear decision boundaries and perform the required transformations sequentially in each hidden layer. Thus, the nonlinearity enables the approximation of virtually any function, allowing for a universal approximation.

Thus, for example, in image recognition it is possible for the recognition in a neural network to be divided into individual hidden layers so that, for example, the simple boundaries of an object are recognized in the first hidden layer of neurons, its structures in the next layer, and only in the subsequent layers the composite whole. Similarly, in the case of language models (LLMs), simpler word dependencies are recognised in the first hidden layers and only in the next deeper layers are more complex language patterns recognised. Experimental research "suggests that LLMs tend to effectively categorize easy tasks, suggesting that these concepts are learned in the first few initial layers. In contrast, complex tasks may not be recognized (if at all) until deeper layers, and LLMs of the same size show largely consistent results across datasets in terms of concept depth." (Jin et al. 2025)

As stated above, the determinism of the symbolic approach is fully predictable, and although some possible shortcomings in technical implementation must be taken into account, the implementation of the algorithm within the symbolic approach involves executing previously known logical steps (rules) that lead to the desired result. In this case, the computing system progresses through a series of consecutive states as it executes the individual steps of the algorithm. According to these steps, the algorithm is built and implemented sequentially, rule by rule. In the case of the neural approach, the weights at each node are modified sequentially during the learning process, and it is not clear in advance what their distribution will be and when they will be established so that the model is successful in a particular learned capability. Here, we must acknowledge that the determinism of a neural network is challenging to predict, and this difficulty increases as the complexity of the neural network increases. "[U]nderstanding what is happening inside of a model regarding domain expertise is highly nontrivial, when it is even possible." (Kent 2023)

Such doubts also relate to other issues, such as whether there is a theoretical limit to the capabilities that models can acquire and how much those capabilities are determined by the data. This uncertainty has two possible causes which exhibit certain symmetries.

(1) There is no analytical way to infer the behaviour of the model as a whole from knowledge of the behaviour of each and every individual neuron at the micro-level, i.e., for example, what macro-level abilities the model will possess. At the same time, there is also no analytical way to determine the micro-level behaviour of individual neurons from knowledge of the model's macro-level abilities. Thus, we might venture to say that the micro and macro levels are symmetrically irreducible.

(2) There is no analytical way to unambiguously predict the model's acquired behaviour given the data on which the model has been trained – hypothetically, this means that the model may not detect all relevant patterns hidden in the data, just as it may learn to detect patterns in the data of which we are unaware. At the same time, there is also no analytical way to determine from knowledge of the model's acquired abilities on what data the generalisation was trained. However, this does not mean that the model's ability to determine whether it had specific data available during learning is impossible; rather, it means that one cannot say from what set of data a given capability or generalisation was derived. The model's ability to generalise in a certain way and the possible generalisations of patterns hidden in the data are also mutually irreducible.

Thus, it is very difficult and probably impossible, given the complexity and the system's dynamic mentioned above, to predict *ex-ante* the model's macro-behaviour, i.e. the abilities that the model will have and that it could learn from a given dataset, just as it is impossible to predict the model's micro-properties, i.e. which neurons in which layers will be activated when these macro-features are detected. In this context, an analogy is offered with cellular automata, whose behaviour also cannot be reduced by other analytical means, and predicting the state of an automaton after *n* steps cannot be done except *by simulating* it. In the context of neural network models, the term "simulation" can be confusing and is better understood as *realisation* or *actualisation*, meaning that it is not possible to infer system properties except by making them occur. Mark Bedau uses the phrase "underivability except by simulation" in the context of cellular automata, and in his view, even a hypothetical Laplace supercalculator would have to "simulate" to arrive at a given state of system properties. "Underivability without simulation is a purely formal notion concerning the existence and nonexistence of certain kinds of derivations of macrostates from a system's underlying dynamics." (Bedau 1997, 379) Such systems, Bedau calls "weakly emergent" (Bedau 1997, 378). Hypothetically, one could assume that a similar principle might apply in the context of DNNs. The complexity of the system dynamics of neuronal activity at the micro-level does not allow for the derivation of the macrostates of acquired system properties other than by realising them in a gradual fine-tuning in the learning process. The realisation of these emergent abilities depends on certain model parameters, and it is now necessary to consider how these phenomena might depend on scaling.

## 4. Scaling and Power Laws

The first DNN models were developed based on the intuitive assumption that as the model parameters increase, their capabilities and power will also increase. It was not only intuition but also experimental practice that confirmed the theoretical analogies related to biological systems. "We believe that deep networks excel because they exploit a particular form of compositionality in which features in one layer are combined in many different ways to create more abstract features in the next layer. For tasks like perception, this kind of compositionality works very well and there is strong evidence that it is used by biological perceptual systems." (Bengio et al. 2021, 60)

Experimental results in the present decade "strongly suggest that larger models will continue to perform better, and will also be much more sample efficient than has been previously appreciated. Big models may be more important than big data." (Kaplan et al. 2020) Scaling laws in the context of LLMs describe how the performance of these models evolves with changes in key factors such as their size (the number of model parameters N), the amount of data they are trained on (the size of the dataset D), and the computational resources utilised during training (the amount of compute C). These laws often manifest as predictable patterns, frequently exhibiting a power-law relationship, where performance improvements are proportional to the scale of these factors raised to a certain power.

For a specific period, it was possible to predictably improve performance on tasks such as text prediction and grammatical accuracy. For example, models such as GPT-3 and PaLM showed smooth improvement in next-word prediction as they scaled and obeyed empirically derived power-laws. Google Research introduced the Pathways Language Model (PaLM), a 540-billion-parameter model, and "pushing the limits of model scale enables breakthrough few-shot performance of PaLM across a variety of natural language processing, reasoning, and code tasks." (Narang and Chowdhery 2022) OpenAI demonstrated that "transformer-based models' capabilities improved in predictable ways as they increased the model size, dataset size, and amount of compute used for training. When all three factors were scaled in concert, model performance followed a smooth power-law curve." (Garg 2024) Similarly, Kaplan states that "Performance has a power-law relationship with each of the three scale factors N, D, C when not bottlenecked by the other two, with trends spanning more than six orders of magnitude. We observe no signs of deviation from these trends on the upper end, though performance must flatten out eventually before reaching zero loss." (Kaplan et al. 2020, 3)

This quantitative framework is essential for comprehending the impact of resource allocation on the capabilities of LLMs. However, not all is clear. For example, empirical research has revealed that the improvement in performance observed depends strongly on scale and weakly on model shape, as well as other architectural hyperparameters, including depth versus width (Kaplan et al. 2020). These empirical results suggest that DNNs exhibit some adaptivity with respect to their architecture. If strongly dependent on the range of parameters and weakly dependent on the layout, it seems that models are able to adapt to a given architecture if they have sufficient resources to do so, replacing, for example, depth (i.e. the number of hidden levels) with width (i.e. the number of neurons in a level) and vice versa. However, one can reasonably argue that such adaptability must have limits given that, as we have seen, nonlinearity is essential for obtaining more complex and abstract capabilities, and these should not be achievable with a more extreme ratio of model width to depth. It would therefore be interesting to experimentally establish bounds on, for example, the width-depth dependence of given models, or more generally, the relationship between scale and network design. Berti et al., in their survey of the emergent abilities of LLMs, say that "The interplay between model scaling, architectural design, and the onset of novel capabilities remains incompletely understood, necessitating further investigation." (Berti et al. 2025)

In addition to the question of the emergence of abilities depending on scale and design, there is also the question of whether there are some fundamental theoretical constraints, perhaps illuminated by perspectives such as dynamic systems theory, that limit the capabilities achievable through mere scaling. We cannot say if LLM scaling will continue its exponential ascent, or if the field is nearing an inflexion point characterised by a plateau or even a decline in the efficacy of pure scaling. In 2024, many theorists compared the changes in previous successful developments of LLMs and described the recent situation as a plateau in the developmental graph. For example, the transition from GPT-3 to GPT-4 achieved much smaller improvements than the earlier dramatic leap from GPT-2 to GPT-3. GPT-4 did not show significant consistent improvements in capabilities in some areas (e.g., coding), despite consuming significantly more resources than its predecessors. Whether we have thus reached a kind of optimal scale beyond which the performance curve of the models must flatten, or whether scaling can instead achieve further emergent capabilities in individual models, remains a debated and open question (Wei et al. 2022). Many theorists believe that scaling is not enough, but rather that we need to explore alternative network arrangements and utilise architectures for DNNs other than the successful Transformer (Vaswani et al. 2017).

5. **Breakthroughs, phase change, emergence**

Current discussions about the emergent properties of LLMs have been sparked by observations of surprising behaviour in individual models before they achieve these new, remarkable abilities. The intuitively expected gradual increase in performance of the observed abilities was often disrupted by some discontinuity and an unexpected, sudden increase in performance. "Though performance is predictable at a general level, performance on a specific task can sometimes emerge quite unpredictably and abruptly at scale." (Ganguli et al. 2022, 4) Given this, the models are, in a general sense, highly predictable because they improve their capabilities as a function of scale, and at the same time, highly unpredictable because it is impossible to determine in advance when and what capabilities will emerge(Ganguli et al. 2022, 2). The emergent abilities of LLMs refer to those features whose existence depends on the computational scales of the model, and such features appear unpredictably and suddenly, whether the models are explicitly trained on them or not. Such abrupt improvements in a model are described by experimenters as "discontinuous dynamics" and "phase transitions" (Chen et al. 2024), "breakthroughs" (Ganguli et al. 2022; Srivastava et al. 2023), "emergence" (Wei et al. 2022), "breaks" (Caballero et al. 2023), "phase change" (Olsson et al. 2022), and "phase transitions in semantic space" (Marin 2025). The terms have often been used interchangeably in the recent literature (Chen et al. 2024) but the term "emergence" appears to play a distinct role, as there are attempts to define it in relation to LLMs.

For example, in an influential and widely discussed paper on the emergent abilities of LLMs, Wei et al. define these abilities with respect to small and large-scale models: "An ability is emergent if it is not present in smaller models but is present in larger models." (Wei et al. 2022) The reason for this definition of an emergent ability lies in the possibility of prediction by extrapolating a scaling law (i.e. consistent performance improvements) from small-scale models(Wei et al. 2022). In a wide discussion of their article on OpenReview, they add "... there are plenty of abilities that are not emergent. For example, many abilities arise smoothly as a result of model scale, some abilities are inversely correlated with model scale, and yet other abilities never achieve performance above random for any model." (Wei et al. 2023) From this perspective, a smoothly developing ability is not emergent; only those abilities for which improvement cannot be predicted by extrapolating a scaling law are considered emergent, i.e., they undergo a sudden change in improvement.

It seems to me that such a distinction cannot be satisfactory when considering the dynamic complexity of the whole. Similarly, in the discussion on OpenReview, Joshua Kimrey notes in one critical comment: "How can I tell whether I am observing an emergent ability or something that is simply a surprising model behaviour?" (Wei et al. 2023) The problem with this kind of definition is that it is too phenomenological, as it does not account for the internal mechanisms and creates the false impression that only unpredictable abilities are emergent, i.e., those that suddenly improve with increasing scale, as described by the shifts in graphs. "We call a prompted task emergent when it unpredictably surges from random performance to above-random at a specific scale threshold." (Wei and Yi 2022)

It is true that many emergent properties in different fields of science are often associated with phase transitions and, of course, this can generally be a good sign of emergent behaviour. In these cases, however, the discontinuous dynamics do not depend on scale but rather on certain critical system parameters such as temperature, which measures the average kinetic energy of the particles in a substance. Although the complexity of a huge number of interacting entities is a fundamental prerequisite for these transitions, it is not a scaling parameter. The fact that similar phenomena appear in the context of LLMs in connection with scaling does not yet imply that scaling is the only cause of emergence. Emergent behaviours are not just a byproduct of scale but are deeply tied to the learning dynamics of neural networks and the internal dynamics of entities and their relations inside the whole system. Hypothetically, we can imagine a situation in which we would only use sufficiently large models in which all acquired abilities would occur easily, so that we would not know their success in small models. As a result, we would also be unable to define their emergent properties in relation to scale. However, this does not imply that such abilities do not arise or have no causes.

Other ways of defining emergent phenomena in LLMs are no better. For example, Rogers and Luccioni mention three other definitions that appear in the recent literature, in addition to the model scale dependency mentioned above: (1) "A property that a model exhibits despite not being explicitly trained for it."; (2) "A property that the model learned from the pre-training data."; (3) "Their sharpness, transitioning seemingly instantaneously from not present to present, and their unpredictability." (Rogers and Luccioni 2023, 6) Each of these aspects expresses a specific part of emergent behaviour, or rather, the consequences of processes occurring in dynamic nonlinear systems that lead to phenomena manifesting themselves in this way. Therefore, it is misleading to understand these partial aspects as possible *competing* definitions, as they are only accompanying features or aspects of general emergent behaviour. One of these aspects of emergent behaviour was discovered by researchers at OpenAI, which in this case was not dependent on scaling but only on the DL process. This kind of behaviour they called Grokking or Delayed Generalisation.

## 6. Grokking or Delayed Generalisation

The term "Grokking" was first used in the context of artificial intelligence and machine learning in 2022 to describe the phenomenon where a model suddenly showed a significant improvement in performance after a period of apparent stagnation during training. Researches from OpenAI "show that, long after severely overfitting, validation accuracy sometimes suddenly begins to increase from chance level toward perfect generalisation. We call this phenomenon 'grokking.'" (Power et al. 2022) The term "grokking" is cleverly borrowed from Robert Heinlein's novel "Stranger in a Strange Land" (1961 [1961]), where "'to grok' means to understand so thoroughly that the observer becomes part of the observed" (pp. 285-286). In a figurative sense, this means here that the neural network initially only remembers training examples without really understanding the patterns but after prolonged training, it suddenly shows a dramatic improvement in its ability to generalise even to previously unknown data. Power et al. believe that the hypothesis that "grokking may only happen after the network's parameters are in flatter regions of the loss landscape" (Power et al. 2022) needs to be examined more closely.

Grokking thus violates our common assumptions and intuitions associated with machine learning, where we expect gradual and progressive improvement. It is surprising that such small and seemingly insignificant changes, persisting long after the interpolation threshold is reached, can lead to such a sudden improvement in delayed generalisation, i.e. a sudden qualitative increase in generalisation ability. Arising from small and gradual quantitative changes, such a sudden qualitative jump cannot be expected and calculated in advance. In general, such a class of phenomena is not only fascinating in the way it puts our intuition at odds with reality but also should be sufficiently compelling for those interpretations of the abilities of DNNs that reduce them to mere "stochastic parrots", the blind repetition of the remembered. In the case of grokking, one can well distinguish the stages in the model's abilities when the model only simply applies what is remembered and the stage when the model suddenly acquires the ability to understand the general pattern and the ability to apply it to new data that was not previously available to the model during the learning phase. Thus, this is similarly unexpected behaviour to previous scaling cases and clearly these sudden changes in model capabilities are the result of more causes than just parameter scaling. Again, the analogies with emergent phenomena in other domains of reality cannot be ignored and this is another reason to seriously consider these sudden qualitative changes as phase transitions from mere memorisation to understanding. Additionally, recent research by Anthropic (Anthropic 2023a; 2023b; 2025; Kahn 2025) suggests that contemporary models exhibit higher abstraction and generalisation capabilities than less powerful, simpler models. Unlike early grokking studies on small models, recent research leads to "the first empirical evidence of grokking in large-scale LLM pretraining, showing that generalisation emerges asynchronously across domains—often well after training loss has converged." (Li et al. 2025) The authors try to demystify grokking's "emergence of generalisation" by investigating the dynamics of LLMs' internal states. They have found that the pathways of training samples evolve from random to more structured and shareable among samples. "These indicate a memorization-to-generalization 'knowledge digestion', providing a mechanistic explanation of the delayed generalization." (Li et al. 2025)

If grokking or delayed generalisation appears across domains and evolves similarly, then it has the nature of a regularity or perhaps a law which is dependent on the neural network structure and learning processes. However, trying to demystify 'emergence' via a mechanistic explanation is not a satisfactory goal. Firstly, 'emergence' is not mystical in the sense of being an unnatural mysterious process which causes the appearance of new things or properties in the world. Secondly, a 'mechanistic explanation' may be an attempt to detect the internal activities of parts of the model in greater detail but, given its complexity, there is no satisfactory way how, in principle and even more so in practice, to deduce the specific system's generalisation from the detailed description of each neuron's activity; and this is precisely an emergence, a natural process: how, in the complex of relations between known identical entities with similar properties, something new emerges – a new property or ability that is unattainable for each individual entity, even though it is also unattainable without the cooperative addition of all entities. It is neither magical in a mysterious way nor mechanical in a reductionist way.

## 7. Metrics, Pre-training Loss and In-Context Learning (ICL)

The "mirage" hypothesis was proposed by Schaeffer, Miranda, and Koyejo (Schaeffer et al. 2023) as an alternative to the scale hypothesis (Wei et al. 2022), aiming to provide evidence that emergent phenomena may not be fundamental properties of scaling AI models but rather byproducts of measurement and metric choices. The scale hypothesis posits that emergent capabilities appear suddenly, transitioning from a non-existent to an existent state seemingly instantaneously, and are unpredictable, appearing unpredictably based on the model scales. The mirage hypothesis, however, considers that emergent capabilities can be illusory because the choice of metrics in graphs influences their course. Schaeffer et al. claim "that specifically, nonlinear or discontinuous metrics produce seemingly emergent abilities, whereas linear or continuous metrics produce smooth, continuous, predictable changes in model performance." (Schaeffer et al. 2023) This means that for "such discrete metrics, the jump is an expected behaviour and provides evidence against labelling this as 'emergence,' i.e., an unpredictable jump." (Schaeffer et al. 2023) Schaeffer et al. do not say that we cannot use these types of metrics, or that it is better to use continuous metrics, but rather, if unpredictability as one of the main characteristics of emergence is the reason for calling behaviour emergent then with the change of metrics to continuous these behaviours can be more predictable and emergence is a mirage.

On the other hand, the use of discrete metrics is a standard way of displaying the development of monitored properties in a graph within a reasonable space, and this does not mean that the use of continuous metrics would alter the fact of obtaining a given property. Experimentally, testing the development in a much larger case, using gradually scaled models, would also be necessary. Finally, a somewhat unfortunate discussion could arise if the emergence of an emergent property were determined by the sufficient steepness of the curve in the graph.

Moreover, Du et al. (2025) have proposed an alternative approach to defining emergence in LLMs, based on pre-training loss thresholds. They propose a shift in analytical focus from model size or training compute to pre-training loss as the primary independent variable for studying emergent abilities. They observe that "a model exhibits emergent abilities on certain tasks, regardless of the continuity of metrics, when its pre-training loss falls below a specific threshold. Before reaching this threshold, its performance remains at the level of random guessing." (Du et al. 2025) Du et al. suggest that the "sharpness" of performance improvement is linked to the model crossing a critical learning threshold, as indicated by its pre-training loss, rather than being solely an artefact of the metric applied. "We conclude that emergent abilities of language models occur when the pre-training loss reaches a certain tipping point, and continuous metrics cannot eliminate the observed tipping point." (Du et al. 2025, 8) They propose a redefinition of emergent abilities based on these findings: "an ability is emergent if it is not present in language models with higher pre-training loss, but is present in language models with lower pre-training loss." (Du et al. 2025, 8)

The arguments presented by Du et al. shift the debate from merely whether a metric is continuous to how that metric is interpreted in the context of task-relevant baselines. While Schaeffer et al. attempt to show only that qualitative jumps depend solely on the continuity and discontinuity of representations depending on metrics, Du et al. emphasise that even continuous representations can contain a practically significant "jump" if this jump represents a transition from performance below the baseline level to performance above the baseline level. This implies that the "mirage" question is not solely about the metric's mathematical form but also about its semantic interpretation in relation to actual task mastery. This points away from the monolithic "all emergence is a mirage" or "no emergence is a mirage" stances, towards a more nuanced understanding where the reality of observed performance jumps depends on the specific task, the model's learning trajectory

Beyond the debate on metrics and pre-training loss, another significant line of critique and alternative explanation for emergent abilities revolves around the powerful capabilities of in-context learning (ICL). Lu et al. concluded that "in-context learning" (i.e., few-shot prompting) is essential for emergent functional abilities: "Our findings suggest that purported emergent abilities are not truly emergent, but result from a combination of in-context learning, model memory, and linguistic knowledge." (Lu et al. 2024) The authors claim that their research proved two hypotheses: "a) That the emergence of all previously-observed functional linguistic abilities is a consequence of ICL; and b) That the abilities which present themselves in instruction-tuned LLMs are more likely to be indicative of instruction-tuning resulting in implicit ICL, rather than the emergence of functional linguistic abilities." (Lu et al. 2024) In short, functional linguistic abilities are not truly emergent but are merely a consequence of ICL. It means that the obtained abilities are rather illusory and their true character is different, non-emergent. However, it is unclear under what conditions a significant improvement in model behaviour would be considered 'truly' emergent, given that the contribution of ICL improves behaviour in a non-emergent way. If I understand the authors correctly, they want to assume that a significant improvement in model behaviour could be understood as truly emergent if it resulted solely from the internal states of the model, without the contribution of ICL.

However, I am concerned that this would lead to an inappropriate conception of neural network models. The trained model is defined by the fixed architecture of connections between neurons in the input, hidden, and output layers, as well as the stored scales of vectors (a matrix of weights and biases). Yet without the dynamic transformation of information over time, there is nothing more than a vast amount of meaningless numbers. It is a trivial fact that the point of LLMs is the transformation of inputs to outputs. From this perspective, inputs and outputs are fundamental integral parts of DNNs, not marginal external components. Now, can I say that there are special kinds of inputs, such as ICL or Chain-of-Thought (CoT) prompting, which are so unique that the model's obtained abilities are caused solely by these types of inputs and not by the model itself? I think not, for the following reason.

It begins with the question of whether it is possible, on a micro-level basis, to differentiate between reasonable and meaningless inputs. If we lack access to a macro-level source of the prompt, we can only see the "On" and "Off" patterns of neuron states in the input layer without knowing about the source (e.g., semantic) content. From this micro-level basis, all states (patterns of inputs) are equally ontologically justified. If inputs are an integral part of the network, as we see above, and if we cannot differentiate the content of inputs on the micro-level, then all activity of the neural network, including inputs and outputs, is a unique consequence of the model as a whole. This means there is no reliable mechanism for recognising on a micro-level basis which part of the inputs are instructions, ICL, or CoT prompting, or the content of the task. From this ontological perspective, the behaviour of the model is emergent and is not dependent on the form of inputs but only on the model's structure and the inner mechanism of interactions between its main components, neurons. Sudden changes and subsequent success in some models' abilities are emergent, not due to scale, pretraining loss, ICL, or CoT, but because DNNs alone are a new field of complexity that leads as a whole to emergent behaviour. In complexities such as DNNs there is not only one way and one cause which leads to the emergent event, ability or behaviour. All of this would be a misunderstanding of emergence, and the very definition of it in the context of LLMs still remains fluid and contested if we do not define it with regard to the inner dynamic transformation of information inside DNN processes.

## 8. Emergent properties in DNNs

Recently, many authors have argued that the modelling of DNNs has shifted from a scientific engineering discipline to a science of complex systems in which the behaviour of these models must be studied because they exhibit emergent properties. Emergent behaviours in LMs are discovered, not designed (Holtzman et al. 2023; Bubeck et al. 2023; Wei et al. 2022; Teehan et al. 2022).

One of the first experimental findings on emergent linguistic structure, which contradicts Bender's "Stochastic Parrots" critique (Bender et al. 2021), demonstrates that modern deep contextual language models can learn significant aspects of language structure without any explicit supervision (Manning et al. 2020). The fact that the model simply tries to predict the masked word in a given context during the learning phase does not mean that it only learns the probabilities of the possible next word in that context and then only behaves like a parrot. In experimental research with the BERT model trained via self-supervision on word prediction tasks, the authors have demonstrated the surprising extent to which the model implicitly learns to recover the rich latent structure of human language. That such rich information emerges through self-supervision is surprising and exciting(Manning et al. 2020)

However, the concept of system properties emerging in connection with DNNs is not new. The current Nobel laureate in physics, John J. Hopfield, tested the theoretical possibility of emergent (collective) properties in integrated circuits as early as 1982, inspired by his research in neurobiology on the molecular dynamics of molecules. He extended existing knowledge of circuits of a few neurons that exhibited elementary biological behaviour, and asked whether the ability of large collections of neurons to perform 'computational' tasks might be partly a spontaneous collective consequence of the existence of large numbers of interacting simple neurons. He suggested that, for example, "the stability of memories, the construction of categories of generalization, or time-sequential memory" might also be emergent properties of collective origin, and he suggested that such important properties arise spontaneously. (Hopfield 1982, 2554)

Hopfield's belief in the spontaneous emergence of collective emergent properties of the whole foreshadowed current thinking about the properties LLMs supposedly possess without being deliberately trained for them in advance. As we have shown above, in recent, much more complex DNN models, there are qualitative leaps in the ability to solve complex tasks when certain thresholds are exceeded. In this context, attention is being paid to advanced functions, such as multilevel reasoning, mathematical proof, and contextual understanding, that emerge spontaneously without explicit training on these tasks. These investigations aim to uncover and understand the relationship between the different parameters of these models and the emergence of these abilities. As shown above, the reason lies primarily in the complexity of these nonlinear dynamical systems and a specific transformation of information inside them. Interestingly, there are analogies in other natural sciences, not only physics, chemistry and biology, but also in all fields of complex dynamical systems. Now it seems evident that we not only discover new fields of dynamical nonlinearities in nature through DNNs, but in this case we also participate in their creation.

This research has a remarkable analogy with the physical law of ideal gases. As some authors emphasise, "our scaling relations go beyond mere observation to provide a predictive framework. One might interpret these relations as analogues of the ideal gas law, which relates the macroscopic properties of a gas in a universal way, independent of most of the details of its microscopic constituents. [...] It would also be exciting to find a theoretical framework from which the scaling relations can be derived: a 'statistical mechanics' underlying the 'thermodynamics' we have observed." (Kaplan et al. 2020) The analogy between statistical mechanics as a low-level theory and thermodynamics as a higher-level theory is also attractive to other authors: "Our work can be viewed as a step towards a statistical physics of deep learning, connecting the 'microphysics' of low-level network dynamics with the 'thermodynamics' of high-level model behavior." (Z. Liu et al. 2022) Similarly, further authors find their research analogous to the relationship between chemistry and biology: "A lack of clarity on what models do holds us back, as if we were studying organic chemistry without knowledge of biology." (Holtzman et al. 2023) Hopfield was also inspired by certain physical analogies that could support the implications of the collective contributions of neurons within the whole: "In physical systems made from a large number of simple elements, interactions among large numbers of elementary components yield collective phenomena such as the stable magnetic orientations and domains in a magnetic system or the vortex patterns in fluid flow. Do analogous collective phenomena in a system of simple interacting neurons have useful 'computational' correlates?" (Hopfield 1982, 2554)

In all cases, the authors emphasise the difficulty of moving from the micro-behaviour of the constituents of the systems in question to the macro-behaviour at the level of the whole. In the analogies between computational science and other areas of the special sciences mentioned above, the authors emphasise that it is possible to define regularities at macro levels without being clear how they are generated by entities at lower levels that contribute to the whole through their more elementary activity. For example, in physics, within thermodynamics, macroscopic laws for phenomenological quantities such as pressure, temperature, and volume were first formulated without seeking causes and explanations for these laws at lower molecular or atomic levels, and only later did it become apparent that these laws were derivable from statistical mechanics describing the behaviour of many particles at lower levels. Thus, for this reason, the authors assume by analogy the existence of a deeper theoretical framework in computational theory from which, so far, only phenomenologically observed regularities in model behaviour would emerge as consequences of deeper (perhaps statistical) regularities.

Such parallels between physics and computational theory are certainly interesting but the success of such explanations presupposes a strong reductionism. Whether one can actually deduce the laws of thermodynamics from statistical mechanics is still a debated question because there are still some fundamental "parameters" that cannot be theoretically deduced from the micro-level but require experiments and approximations, i.e., substantial consideration of the macro-level, to make this so-called "derivation" possible. For many, this fact is a decisive factor in the unattainability of such a reduction (Sklar 1999; Bishop and Atmanspacher 2006; Batterman 2011; Bishop and Ellis 2020) and some consider this reductionist assumption outright a "famous reductionist legend" (Bishop and Ellis 2020).

If we assume that the aforementioned parallel does not apply only between physics and computational science, but instead applies more generally to any relationship between systemic laws (at the macro level) and the laws of behaviour of its constituents (at the micro level), then we can reasonably assume that in many different areas of reality the principle holds whereby systemic consequences cannot be clearly deduced from the behaviour of constituents at the micro level. In reasoning about the properties of the DNNs, the systemic properties of the whole, i.e., the capabilities of a given model acquired in the DL process, cannot be deduced from the behaviour of individual neurons in the network. In other words, the properties of DNNs are emergent properties with respect to the level of properties of individual neurons, and that by this principle DNNs are functionally similar not only to their biological model (the neuronal structure of the brain) but also that they resemble in their behaviour all other complex structures, which consist of many identical constituents in interconnections that interact in a nonlinear way to give rise to a non-trivial resultant behaviour of the whole. This emergent principle has been successfully exploited to provide LLMs with abilities, such as pattern recognition, understanding meaning and natural language, and perhaps also intentionality, which had hitherto been, at least from some philosophical positions, attributed exclusively to the brain as a biological organ. In this context, it is worth remembering that attempts to explain the existence of mind and consciousness in the material world substantively were soon replaced by approaches that discuss the existence of consciousness as an emergent property of the brain's complexity (e.g. Churchland 1989; Sperry 1980; Dennett 1991; Searle 1992; Chalmers 1996). The extent to which this principle can be used in the context of DNNs to acquire additional abilities, and perhaps even abilities that we cannot yet imagine, is still an open question.

**Conclusions**

Recent discussions about the emergent properties of LLMs demonstrate that although we are creating model phenomena that have counterparts in organic neural structures, their behaviour, based on the complexity of the entities involved, exceeds our predictions and is probably subject to some general natural principle of mutual cooperative dynamics. It is understandable that phenomena we create but do not fully understand or control may raise various concerns. In this paper, the focus has been solely on possible epistemological concerns about the explainability and predictability of such phenomena, which, however, stem primarily from exaggerated reductionist ideas. It has been demonstrated that the comprehensible nature of these phenomena also encompasses a certain incomprehensibility, albeit in a reductionist sense.

As for the discussion of the causes of such phenomena, the scaling of parameters is essential for the appearance and improvement of LLMs' properties but it is not the only cause of the emergence of these properties. The existence of these properties or abilities also depends on the impact of other parameters, such as the network's architecture or design, data and DL processes, and types of prompting, including ICL or CoT. Thus, while LLMs' emergent properties are not apparent consequences of discontinuous, standard imaging metrics, they cannot be defined only phenomenologically in terms of scaling models. Intuitively, larger models undoubtedly have more capacity to possess such capabilities but the existence of such capabilities in larger models and their absence in smaller ones is simply a consequence of their dependence on internal entities, their relations, and their transformation within the neural network. It can be assumed that certain axioms, ontological in nature, which are independent of the specific nature of entities and the dynamics of the relations between them, must be satisfied for emergent properties to appear in general.

DNNs, as highly complex systems, are highly sensitive to seemingly subtle changes, which allows them to acquire system-level capabilities, i.e. macro-properties that are co-created by the elementary operations of the fundamental micro-entities of the network, i.e. neurons that do not possess such system properties themselves. Since this satisfies the universal emergent principle (Havlík 2022), the capabilities thus acquired by the models must be emergent. As we can note in complexities such as DNNs, there is not only one way and one cause which leads to the emergent behaviour of the whole. It would be a misunderstanding of emergence if we were to think that we need to identify a special kind of cause that leads to an emergent property. The definition of the term "emergence" in the context of LLMs may remain unclear and controversial unless it is defined in relation to the internal dynamic transformation of information within DNNs.

## Acknowledgements

This work has been funded by a grant from the Programme Johannes Amos Comenius under the Ministry of Education, Youth and Sports of the Czech Republic, CZ.02.01.01/00/23_025/0008711.

## References

Anthropic. 2023a. 'Towards Monosemanticity: Decomposing Language Models With Dictionary Learning'. October 4. https://www.anthropic.com/news/towards-monosemanticity-decomposing-language-models-with-dictionary-learning.

Anthropic. 2023b. 'Decomposing Language Models Into Understandable Components'. October 5. https://www.anthropic.com/news/decomposing-language-models-into-understandable-components.

Anthropic. 2025. 'On the Biology of a Large Language Model'. Transformer Circuits. https://transformer-circuits.pub/2025/attribution-graphs/biology.html.

Antonion, Jackob, Annie Wang, Maziar Raissi, and Ribana Joshie. 2024. 'Nondeterministic Features in Deep Neural Network Design, Training and Inference'. *Mathematical Modeling and Algorithm Application* 2 (3): 5–9. https://doi.org/10.54097/aze6m665.

Batterman, Robert W. 2011. 'Emergence, Singularities, and Symmetry Breaking'. *Foundations of Physics* 41 (6): 1031–50. https://doi.org/10.1007/s10701-010-9493-4.

Bedau, Mark A. 1997. 'Weak Emergence'. *Philosophical Perspectives* 11: 375–99. https://doi.org/10.1111/0029-4624.31.s11.17.

Bender, Emily M., Timnit Gebru, Angelina McMillan-Major, and Shmargaret Shmitchell. 2021. 'On the Dangers of Stochastic Parrots: Can Language Models Be Too Big? 🦜'. *Proceedings of the 2021 ACM Conference on Fairness, Accountability, and Transparency* (New York, NY, USA), FAccT '21, March 1, 610–23. https://doi.org/10.1145/3442188.3445922.

Bengio, Yoshua, Yann Lecun, and Geoffrey Hinton. 2021. 'Deep Learning for AI'. *Communications of the ACM* 64 (7): 58–65. https://doi.org/10.1145/3448250.

Berti, Leonardo, Flavio Giorgi, and Gjergji Kasneci. 2025. 'Emergent Abilities in Large Language Models: A Survey'. arXiv:2503.05788. Preprint, arXiv, March 14. https://doi.org/10.48550/arXiv.2503.05788.

Bishop, Robert C., and Harald Atmanspacher. 2006. 'Contextual Emergence in the Description of Properties'. *Foundations of Physics* 36 (12): 1753–77. https://doi.org/10.1007/s10701-006-9082-8.

Bishop, Robert C., and George F. R. Ellis. 2020. 'Contextual Emergence of Physical Properties'. *Foundations of Physics* 50 (5): 481–510. https://doi.org/10.1007/s10701-020-00333-9.

Bubeck, Sébastien, Varun Chandrasekaran, Ronen Eldan, et al. 2023. 'Sparks of Artificial General Intelligence: Early Experiments with GPT-4'. arXiv:2303.12712. Preprint, arXiv, April 13. https://doi.org/10.48550/arXiv.2303.12712.

Caballero, Ethan, Kshitij Gupta, Irina Rish, and David Krueger. 2023. 'Broken Neural Scaling Laws'. arXiv:2210.14891. Preprint, arXiv, July 24. https://doi.org/10.48550/arXiv.2210.14891.

Chalmers, David John. 1996. *The Conscious Mind: In Search of a Fundamental Theory*. Philosophy of Mind Series. Oxford University Press.

Chen, Angelica, Ravid Shwartz-Ziv, Kyunghyun Cho, Matthew L. Leavitt, and Naomi Saphra. 2024. 'Sudden Drops in the Loss: Syntax Acquisition, Phase Transitions, and Simplicity Bias in MLMs'. arXiv:2309.07311. Preprint, arXiv, February 7. https://doi.org/10.48550/arXiv.2309.07311.

Churchland, Paul M. 1989. *A Neurocomputational Perspective: The Nature of Mind and the Structure of Science*. MIT Press.

Dennett, D. C. 1991. *Consciousness Explained*. 1st ed. Little, Brown and Co.

Du, Zhengxiao, Aohan Zeng, Yuxiao Dong, and Jie Tang. 2025. 'Understanding Emergent Abilities of Language Models from the Loss Perspective'. arXiv:2403.15796. Preprint, arXiv, January 15. https://doi.org/10.48550/arXiv.2403.15796.

Ganguli, Deep, Danny Hernandez, Liane Lovitt, et al. 2022. 'Predictability and Surprise in Large Generative Models'. June 21, 1747–64. https://doi.org/10.1145/3531146.3533229.

Garg, Ashu. 2024. 'Has AI Scaling Hit a Limit?' Foundation Capital, November 27. https://foundationcapital.com/has-ai-scaling-hit-a-limit/.

Havlík, Vladimír. 2022. *Hierarchical Emergent Ontology and the Universal Principle of Emergence*. Springer International Publishing. https://doi.org/10.1007/978-3-030-98148-8.

Heaven, Will Douglas Heavenarchive. 2024. 'Large Language Models Can Do Jaw-Dropping Things. But Nobody Knows Exactly Why.' MIT Technology Review, March 4. https://www.technologyreview.com/2024/03/04/1089403/large-language-models-amazing-but-nobody-knows-why/.

Hebb, D. O. 2002. *The Organization of Behavior: A Neuropsychological Theory*. L. Erlbaum Associates.

Heinlein, Robert A. 1961. *Stranger in a Strange Land*. 1st ed. Putnam Publishing Group.

Holtzman, Ari, Peter West, and Luke Zettlemoyer. 2023. 'Generative Models as a Complex Systems Science: How Can We Make Sense of Large Language Model Behavior?' arXiv:2308.00189. Preprint, arXiv, July 31. https://doi.org/10.48550/arXiv.2308.00189.

Hopfield, J J. 1982. 'Neural Networks and Physical Systems with Emergent Collective Computational Abilities.' *Proceedings of the National Academy of Sciences* 79 (8): 2554–58. https://doi.org/10.1073/pnas.79.8.2554.

Jensen, Henrik Jeldtoft. 2022. *Complexity Science: The Study of Emergence*. 1st ed. Cambridge University Press. https://doi.org/10.1017/9781108873710.

Jin, Mingyu, Qinkai Yu, Jingyuan Huang, et al. 2025. 'Exploring Concept Depth: How Large Language Models Acquire Knowledge and Concept at Different Layers?' arXiv:2404.07066. Preprint, arXiv, February 4. https://doi.org/10.48550/arXiv.2404.07066.

Kahn, Jeremy. 2025. 'Anthropic Researchers Make Progress Unpacking AI's "Black Box"'. Fortune, March 27. https://fortune.com/2025/03/27/anthropic-ai-breakthrough-claude-llm-black-box/.

Kaplan, Jared, Sam McCandlish, Tom Henighan, et al. 2020. 'Scaling Laws for Neural Language Models'. arXiv:2001.08361. Preprint, arXiv, January 23. https://doi.org/10.48550/arXiv.2001.08361.

Kent, Jonathan S. 2023. 'Chaos Theory and Adversarial Robustness'. arXiv:2210.13235. Preprint, arXiv, July 5. https://doi.org/10.48550/arXiv.2210.13235.

Li, Ziyue, Chenrui Fan, and Tianyi Zhou. 2025. 'Where to Find Grokking in LLM Pretraining? Monitor Memorization-to-Generalization without Test'. arXiv:2506.21551. Preprint, arXiv, July 3. https://doi.org/10.48550/arXiv.2506.21551.
Liu, Shuhong, Nozomi Akashi, Qingyao Huang, Yasuo Kuniyoshi, and Kohei Nakajima. 2024. 'Exploiting Chaotic Dynamics as Deep Neural Networks'. Version 1. Preprint, arXiv. https://doi.org/10.48550/ARXIV.2406.02580.
Liu, Ziming, Ouail Kitouni, Niklas Nolte, Eric J. Michaud, Max Tegmark, and Mike Williams. 2022. 'Towards Understanding Grokking: An Effective Theory of Representation Learning'. Version 2. Preprint, arXiv. https://doi.org/10.48550/ARXIV.2205.10343.
Lorenz, Edward N. 1972. 'Predictability: Does The Flap of A Butterfly's Wings in Brazil Set Off A Tornado in Texas?' Paper presented at Address at the 139th Annual Meeting of the American Association for the Advancement of Science, Sheraton Park Hotel, Boston, Mass.,. December 29. https://www.scribd.com/document/130949814/Predictability-Does-the-Flap-of-a-Butterfly-s-Wings-in-Brazil-Set-Off-a-Tornado-in-Texas.
Lorenz, Edward N. 2008. *The Essence of Chaos*. Nachdr. The Jessie and John Danz Lectures. Univ. of Washington Press.
Lu, Sheng, Irina Bigoulaeva, Rachneet Sachdeva, Harish Tayyar Madabushi, and Iryna Gurevych. 2024. 'Are Emergent Abilities in Large Language Models Just In-Context Learning?' arXiv:2309.01809. Preprint, arXiv, July 15. https://doi.org/10.48550/arXiv.2309.01809.
Manning, Christopher D., Kevin Clark, John Hewitt, Urvashi Khandelwal, and Omer Levy. 2020. 'Emergent Linguistic Structure in Artificial Neural Networks Trained by Self-Supervision'. *Proceedings of the National Academy of Sciences* 117 (48): 30046–54. https://doi.org/10.1073/pnas.1907367117.
Marin, Javier. 2025. 'A Non-Ergodic Framework for Understanding Emergent Capabilities in Large Language Models'. Version 1. Preprint, arXiv. https://doi.org/10.48550/ARXIV.2501.01638.
McCulloch, Warren S., and Walter Pitts. 1943. 'A Logical Calculus of the Ideas Immanent in Nervous Activity'. *The Bulletin of Mathematical Biophysics* 5 (4): 115–33. https://doi.org/10.1007/BF02478259.
Narang, Sharan, and Aakanksha Chowdhery. 2022. 'Pathways Language Model (PaLM): Scaling to 540 Billion Parameters for Breakthrough Performance'. April 4. https://research.google/blog/pathways-language-model-palm-scaling-to-540-billion-parameters-for-breakthrough-performance/.
Olsson, Catherine, Nelson Elhage, Neel Nanda, et al. 2022. 'In-Context Learning and Induction Heads'. arXiv:2209.11895. Preprint, arXiv, September 24. https://doi.org/10.48550/arXiv.2209.11895.
Power, Alethea, Yuri Burda, Harri Edwards, Igor Babuschkin, and Vedant Misra. 2022. 'Grokking: Generalization Beyond Overfitting on Small Algorithmic Datasets'. arXiv:2201.02177. Preprint, arXiv, January 6. https://doi.org/10.48550/arXiv.2201.02177.
Rogers, Anna, and Alexandra Sasha Luccioni. 2023. 'Position: Key Claims in LLM Research Have a Long Tail of Footnotes'. Version 2. Preprint, arXiv. https://doi.org/10.48550/ARXIV.2308.07120.
Rosenblatt, F. 1958. 'The Perceptron: A Probabilistic Model for Information Storage and Organization in the Brain.' *Psychological Review* 65 (6): 386–408. https://doi.org/10.1037/h0042519.
Schaeffer, Rylan, Brando Miranda, and Sanmi Koyejo. 2023. 'Are Emergent Abilities of Large Language Models a Mirage?' Version 2. Preprint, arXiv. https://doi.org/10.48550/ARXIV.2304.15004.
Searle, John Rogers. 1992. *The Rediscovery of the Mind*. Representation and Mind. MIT press.
Sklar, Lawrence. 1999. 'The Reduction(?) Of Thermodynamics to Statistical Mechanics'. *Philosophical Studies* 95 (1–2): 187–202. https://doi.org/10.1023/A:1004527910768.
Sperry, R.W. 1980. 'Mind-Brain Interaction: Mentalism, Yes; Dualism, No'. *Neuroscience* 5 (2): 195–206. https://doi.org/10.1016/0306-4522(80)90098-6.
Srivastava, Aarohi, Abhinav Rastogi, Abhishek Rao, et al. 2023. 'Beyond the Imitation Game: Quantifying and Extrapolating the Capabilities of Language Models'. arXiv:2206.04615. Preprint, arXiv, June 12. https://doi.org/10.48550/arXiv.2206.04615.19

Summers, Cecilia, and Michael J. Dinneen. 2021. 'Nondeterminism and Instability in Neural Network Optimization'. *Proceedings of the 38th International Conference on Machine Learning*, July 1, 9913–22. https://proceedings.mlr.press/v139/summers21a.html.

Teehan, Ryan, Miruna Clinciu, Oleg Serikov, et al. 2022. 'Emergent Structures and Training Dynamics in Large Language Models'. In *Proceedings of BigScience Episode #5 – Workshop on Challenges & Perspectives in Creating Large Language Models*, edited by Angela Fan, Suzana Ilic, Thomas Wolf, and Matthias Gallé. Association for Computational Linguistics. https://doi.org/10.18653/v1/2022.bigscience-1.11.

Vaswani, Ashish, Noam Shazeer, Niki Parmar, et al. 2017. 'Attention Is All You Need'. arXiv:1706.03762. Preprint, arXiv, December 5. https://doi.org/10.48550/arXiv.1706.03762.

Wei, Jason, Yi Tay, Rishi Bommasani, et al. 2022. 'Emergent Abilities of Large Language Models'. Version 2. Preprint, arXiv. https://doi.org/10.48550/ARXIV.2206.07682.

Wei, Jason, Yi Tay, Rishi Bommasani, et al. 2023. 'Emergent Abilities of Large Language Models'. *Transactions on Machine Learning Research*, June 26. https://openreview.net/forum?id=yzkSU5zdwD.

Wei, Jason, and Tay Yi. 2022. 'Characterizing Emergent Phenomena in Large Language Models'. November 10. https://research.google/blog/characterizing-emergent-phenomena-in-large-language-models/.

Xiao, Guanping, Jun Liu, Zheng Zheng, and Yulei Sui. 2021. 'Nondeterministic Impact of CPU Multithreading on Training Deep Learning Systems'. *2021 IEEE 32nd International Symposium on Software Reliability Engineering (ISSRE)*, October, 557–68. https://doi.org/10.1109/ISSRE52982.2021.00063.